\documentclass[journal]{IEEEtran}
\usepackage[normalem]{ulem}
\usepackage[pagebackref=true,breaklinks=true,colorlinks,bookmarks=false]{hyperref}
\usepackage[pdftex]{graphicx}
\usepackage{array}
\usepackage{xcolor}
\usepackage{multirow}
\usepackage{cite}
\usepackage{amsmath}
\usepackage[caption=false,font=footnotesize]{subfig}
\usepackage{url}
\usepackage{amsfonts}
\usepackage{pifont}

\usepackage{adjustbox}

\newlength\fsdurthree

\begin{document}

\title{Unsupervised haze removal from underwater images}

\author{Praveen Kandula, and~A. N. Rajagopalan, \textit{Senior Member, IEEE} % <-this % stops a space
% \thanks{Mohit Lamba, Kranthi~Kumar~Rachavarapu and Kaushik Mitra are with the Department
% of Electrical Engineering, Indian Institute of Technology Madras, India (e-mail: ee18d009@smail.iitm.ac.in, ee18d004@smail .iitm.ac.in, kmitra@ee.iitm.ac.in).}% <-this % stops a space
% \thanks{This paper has a supplementary downloadable PDF report available at http://ieeexplore.ieee.org., provided by the author. Contact ee18d009@smail.iitm.ac.in for further questions about this work.}
% \thanks{*M. Lamba and K. Kumar have contributed equally to the work.}% <-this % stops a space
}

\markboth{Journal of \LaTeX\ Class Files,~Vol.~14, No.~8, August~2015}%
{Shell \MakeLowercase{\textit{et al.}}: Bare Demo of IEEEtran.cls for IEEE Journals}

\maketitle
\begin{abstract}
Several supervised networks exist that remove haze information from underwater images using paired datasets and pixel-wise loss functions. However, training these networks requires large amounts of paired data which is cumbersome, complex and time-consuming. Also, directly using adversarial and cycle consistency loss functions for unsupervised learning is inaccurate as the underlying mapping from clean to underwater images is one-to-many, resulting in an inaccurate constraint on the cycle consistency loss. To address these issues, we propose a new method to remove haze from underwater images using unpaired data. Our model disentangles haze and content information from underwater images using a Haze Disentanglement Network (HDN). The disentangled content is used by a restoration network to generate a clean image using adversarial losses. The disentangled haze is then used as a guide for underwater image regeneration resulting in a strong constraint on cycle consistency loss and improved performance gains. Different ablation studies show that the haze and content from underwater images are effectively separated. Exhaustive experiments reveal that accurate cycle consistency constraint and the proposed network architecture play an important role in yielding enhanced results. Experiments on UFO-120, UWNet, UWScenes, and UIEB underwater datasets indicate that the results of our method outperform prior art both visually and quantitatively.
\end{abstract}
% \begin{IEEEkeywords}
% Underwater haze removal, haze disentanglement, consistent cycle consistency.
% \end{IEEEkeywords}
\IEEEpeerreviewmaketitle

%%%%%%%%%%%%%%%%
\section{Introduction}
\label{Sec:Introduction}
 
\IEEEPARstart{R}{emoving} haze from underwater images remains a challenging task due to degradation by the participating medium. The low quality of underwater images can be attributed to many factors. Light gets attenuated and scattered by the underwater medium before it reaches the camera lens resulting in colour shifts, low-light, and haze effects in the observed images \cite{uwcauses1,uwcauses2,uwcauses3,uwcauses4}. Additionally, wavelength-based attenuation causes various undesirable colour tones. Specifically, most underwater images have green and blue colour tones as red colour undergoes rapid attenuation. The benefits of removing haze from underwater images include efficient monitoring of coral reefs, marine biology \cite{uwres_uses}, analysis of flora and fauna, in addition to improving high-level computer vision tasks like segmentation and classification of marine animals.

Several works have been proposed in the past few decades for different restoration tasks like deblurring \cite{vasu2018non, mohan2021deep, nimisha2018unsupervised, mohan2019unconstrained, rao2014harnessing, rajagopalan1998optimal, paramanand2011depth, purohit2020region, purohit2019bringing, vasu2017local, nimisha2018generating, rajagopalan2005background}, dehazing \cite{ ancuti2019ntire, purohit2019multilevel, mandal2019local}, inpainting \cite{ suin2021distillation, bhavsar2012range}, enhancement \cite{kandula2023illumination, el2020aim}, super-resolution \cite{ suin2020degradation, rajagopalan2003motion, nimisha2021blind, purohit2021spatially, suresh2007robust, bhavsar2010resolution, purohit2020mixed}, bokeh rendering \cite{purohit2019depth} etc., and many others. Among them, Deep CNN \cite{deepcnn} uses a convolutional neural network (CNN) to remove colour shifts and restores the underlying clean image. WaterGAN \cite{li2017watergan} generates synthetic UW images using terrestrial images and corresponding depth maps. The generated data is then used to train a neural network for removing colour casts in UW images. Simultaneous Enhancement and Super-Resolution (SESR) \cite{sesr} employs an encoder-decoder model for simultaneous enhancement and super-resolution of UW images. Dense GAN  \cite{denseGAN} trains a multi-stage neural network using adversarial and $L_1$ loss functions. Water-Net \cite{uieb} constructs a Underwater Image Enhancement Benchmark (UIEB) datset using an real images and the corresponding references to be the best results from different traditional algorithms. A significant drawback of these methods is the requirement of supervised pairs to train the neural network.

\begin{figure*}[t]
%\newlength\fsdurthree
%\setlength{\fsdurthree}{-1.5mm}
\setlength{\tabcolsep}{1pt}
\scriptsize
\centering
%\hspace{-0.4cm}
\begin{tabular}{ccccccccccc}

% \includegraphics[width=\linewidth]{Images/Cycle_GAN.pdf} \\
% (a) Cycle consistency mismatch in CycleGAN \cite{CycleGAN}. \\
\includegraphics[width=\linewidth]{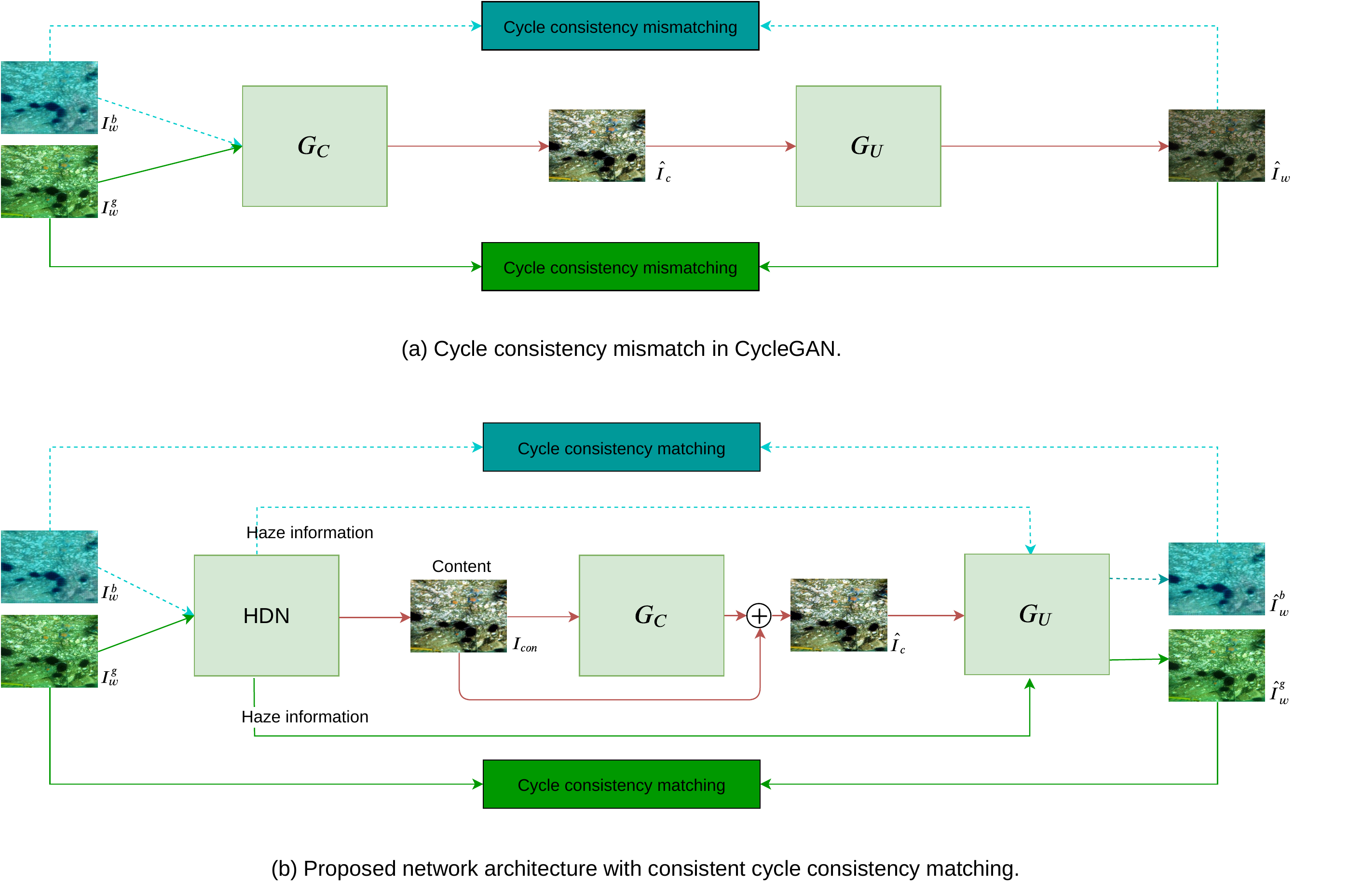} \\
% (b)Proposed network architecture.\\
\end{tabular}\\ %\hspace{-2.3mm}

\caption{(a) Using CycleGAN \cite{CycleGAN} directly for cycle consistency is not sufficient as $G_{U}$ generates underwater images with random haze information that do not match the input underwater images. (b) Haze disentanglement network (HDN) disentangles haze and content information from input images. The disentangled haze is used by $G_{U}$ to regenerate underwater images that match the input images.} 
\label{fig:Network_architecture}
\end{figure*}

\textbf{Unsupervised restoration:} To address the unavailability of supervised pairs, several unsupervised image restoration algorithms have been proposed. Among them, \cite{Cycleincycle} uses two generative adversarial networks (GANs) to transfer images from low-resolution to high-resolution. \cite{nimisha2018unsupervised} makes use of reblurring and gradient losses to restore blurry images. \cite{chellapa} uses KL-divergence to disentangle blur and content information from input images. UEGAN \cite{UEGAN} enhances low-light images using local and global discriminators along with perceptual loss. Similar to \cite{UEGAN}, \cite{enlightengan} additionally utilizes attention maps for image enhancement.

Directly using cycle consistency of \cite{CycleGAN} for haze removal in underwater images is not correct as the underlying mapping from clean to underwater domain is one-to-many. A detailed discussion of cycle consistency mismatch in CycleGAN is given in Sec. \ref{Sec:Proposed_method}. In this paper, we propose an unpaired underwater haze removal algorithm to address the shortcomings of previous works. Our method uses accurate cycle consistency matching loss (see Fig. \ref{fig:Network_architecture} (b)) for removing haze in underwater images. More specifically, we  disentangle haze and content image from underwater images using a haze disentanglement network (HDN). The disentangled content from HDN is used as input to a restoration network ($G_{C}$), which enhances the content information to generate the latent image. The disentangled haze is then used in underwater image regeneration for cycle consistency loss. This loss, combined with adversarial losses on underwater and clean images, successfully removes haze from underwater images. Since the underwater regeneration uses disentangled haze, we employ a correct constraint on cycle consistency matching, unlike vanilla CycleGAN \cite{CycleGAN} (see Fig. \ref{fig:Network_architecture} (a)). During test time, the input underwater image is first passed through the HDN network for content information and then the generated content is passed through $G_{C}$ for the final restored image. 

Our main contributions are listed below:
\begin{itemize}
    % \item To the best of our knowledge, this is the first-ever data-driven attempt to explicitly disentangle haze from underwater images. 
    \item To the best of our knowledge, this is the first learning-based approach to employ correct cycle-consistency loss for unpaired haze removal in underwater images.
    % \item Different ablation studies show that the loss functions and design architecture used in HDN effectively separate haze from underwater images. We verify this by visualizing content and haze information present in underwater images.
    % \item We propose two data augmentation techniques by adding back the disentangled haze in various degrees to underwater image  and transferring the haze to a clean image, for robust training.
    \item Exhaustive experiments on different underwater datasets reveal that accurate cycle consistency matching in conjunction with disentangled content for restoration network gives high-quality dehazing results compared to prior unsupervised methods.
\end{itemize}
% As a second main contribution, we propose two data augmentation techniques for robust training: self and cross augmentation. Given an underwater image, the haze information separated using HDN is added back in various degrees to the input images in self augmentation. As a result, we create various underwater images with different degrees of distortion but for the same underlying scene information. In cross augmentation, the separated haze is transferred to clean images, creating a supervised pair for training. 
% \input{Tex_files/Introduction}
% \input{Tex_files/Related_works}

In our approach, HDN uses feature regularization, feature adversarial, and cyclic losses to decouple haze and content information from input images. Different ablation studies are provided to visualize haze and content information present in underwater images. Exhaustive experiments on different publicly available datasets show that accurate cycle consistency constraint combined with the disentangled content for $G_C$ gives superior results compared to prior unsupervised methods. 
% %%%%%%%%%%%%%%%%%%%%%%%%

% \input{Tex_files/Proposed_method1}

\section{Proposed method}
Fig. \ref{fig:Network_architecture} (b)  outlines the proposed methodology. Our framework has two main parts: haze and content disentanglement from underwater images, and restoration of underwater images using the disentangled information.
\label{Sec:Proposed_method}
\subsection{Haze and content disentanglement}
\label{Sec:Haze_disentanglement}
The objective of HDN is to disentangle haze and content information from an underwater image $(I^{\dagger}_{w})$ using a clean or haze-free image ($I^{\star}_c$). Note that $I^{\dagger}_{w}$ and $I^{\star}_c$ are unpaired images randomly sampled from an underwater and a clean set, respectively. We use haze (haze-free) and underwater (clean) words interchangeably. To achieve these objectives, we use the following loss functions.

\textbf{Feature adversarial loss:} To ensure that $E_{hf}$ extracts only haze-free information from a given input image, we use the following strategy. Since $I^{\star}_c$ is clean, $E_{hf}$ extracts only haze-free information. Let the output of $E_{hf}$ given $I^{\star}_c$ be denoted as $F_{hf_{c}}$ $\in R ^{B\times C \times H \times W}$, where $B, C, H, W$ are, respectively, batch size, number of channels, height and width of feature maps $F_{hf_{c}}$. Similarly, let the output of $E_{hf}$ given $I^{\dagger}_{w}$ be denoted as $F_{hf_{w}}$ $\in R ^{B\times C \times H \times W}$. To ensure that $F_{hf_{w}}$ contains only haze-free information, we use adversarial loss function on feature maps with $F_{hf_{c}}$ as real samples and $F_{hf_{w}}$ as fake samples. The loss function can then be formulated as 
% \begin{equation}
% \begin{split}
%  {L}^{dis}_{{hf}_{adv}}(E_{hf}, E_{h}, D_{adv}) =  \mathbb{E}[\log (D_{adv}(F_{hf_{c}}))] +\\
%  \mathbb{E} [\log(1 - D_{adv}(F_{hf_{w}}))]  
% \end{split}
% \end{equation}
\begin{equation}
\label{Eq: disentangle1}
\begin{split}
 {L}_{{d}_{1}}(E_{hf}, E_{h}, D_{adv}) =  
 \mathbb{E}[\log (D_{adv}(F_{hf_{c}}))] + \\
  \mathbb{E} [\log(1 - D_{adv}(F_{hf_{w}}))]  
\end{split}
\end{equation}

\textbf{Feature regularization loss:} Since $I^{\star}_c$ is clean (or haze-free), $E_{h}$ (the haze encoder) should not respond to it. We use feature regularization loss to ensure this by constraining the output feature maps for all the layers in $E_{h}$ to zero when the input image is $I^{\star}_c$. This loss combined with ${L}_{d{1}}$ (from Eq. \ref{Eq: disentangle1}) ensures that $E_h$ and $E_{hf}$ extract only haze  and haze-free (content) information from input image. The loss function for this objective can be formulated as 
% \begin{equation}
%  L^{dis}_{reg} (E_{h})= ||\sum_{i=1}^{n} F_{h_{c}}||_1
% \end{equation}
\begin{equation}
 L_{{d}_{2}} (E_{h})= \sum_{i=1}^{n} ||F^{i}_{h_{c}}||_1
\end{equation}
where $F_{h_{c}}$ is the output feature map of $E_h$ with $I^{\star}_c$ as input and $i$ denotes intermediate feature maps of $E_h$.

\textbf{Disentangled cyclic losses:}
We use cyclic loss functions to ensure that the decoder ($D$) is sufficiently trained to restore the corresponding input image. Given $I^{\star}_c$, $D$ combines feature maps $F_{hf_{c}}$ and $F_{h_{c}}$ to estimate the corresponding input image. The loss function can be written as 
% \begin{equation}
%  L^{dis}_{hf_{cy}} (E_{hf}, E_{h}, D) = ||D(F_{hf_{c}} + F_{h_{c}} ) - I_c||_1
% \end{equation}
\begin{equation}
\begin{split}
 L_{{d}_{3}} (E_{hf}, E_{h}, D)& = ||D(F_{hf_{c}} + F_{h_{c}} ) - I^{\star}_c||_1 \\
%  & = ||\hat{I}_c - I_c||_1 \\
\end{split}
\end{equation}
 where $F_{hf_{c}}$ and $F_{h_{c}}$ are output feature maps of $E_{hf}$ and $E_{h}$, respectively, for input image $I^{\star}_c$.
Similarly, the cyclic loss function for $I^{\dagger}_{w}$ can be written as 
% \begin{equation}
%  L^{dis}_{h_{cy}} (E_{hf}, E_{h}, D) = ||D(F_{hf_{w}} + F_{h_{w}} ) - I^{\dagger}_{w}||_1
% \end{equation}
\begin{equation}
\begin{split}
 L_{{d}_{4}} (E_{hf}, E_{h}, D) & = ||D(F_{hf_{w}} + F_{h_{w}} ) - I^{\dagger}_{w}||_1 \\
%  & = ||\hat{I}_w - I_w||_1 \\
 \end{split}
\end{equation}
where $F_{hf_{w}}$ and $F_{h_{w}}$ are output feature maps of $E_{hf}$ and $E_{h}$, respectively, for input image $I^{\dagger}_{w}$.

The final loss function to disentangle haze and content information is a combination all the four loss functions and can be written as
% \begin{equation}
% \begin{split}
%     L_{d} (E_{hf}, E_{h}, D, D_{adv}) = \lambda_1 {L}^{dis}_{{hf}_{adv}} + 
%  \lambda_2 L^{dis}_{reg} + \\ \lambda_3 L^{dis}_{hf_{cy}} + \lambda_4 L^{dis}_{h_{cy}}
% \end{split}
% \end{equation}
\begin{equation}
\begin{split}
\label{Eq:disentanglement_losses}
    L_{d} (E_{hf}, E_{h}, D, D_{adv}) = \lambda_1 L_{{d}_{1}} + 
 \lambda_2 L_{{d}_{2}} +  \\ \lambda_3 L_{{d}_{3}} + \lambda_4 L_{{d}_{4}}
\end{split}
\end{equation}
where $\lambda_1$, $\lambda_2$, $\lambda_3$ and $\lambda_4$ are trade-off weights. 

Given an underwater image $I^{\dagger}_w$ as input to the HDN network, the output of haze encoder, $F_{{h}_{w}}$, contains the haze information and the output of haze-free encoder $F_{{hf}_{w}}$ contains content information.
The disentangled content information from HDN is used to generate the content image $I_{con}$ using the following equation
\begin{equation}
\label{Eq. contentimage}
I^{\dagger}_{con} =  D(E_{hf}(I^{\dagger}_w))
\end{equation}
% where $D$ and $E_{hf}$ are decoder and haze-free encoder networks in HDN. 
The haze information and content image separated from an underwater image using HDN network is used by a restoration module to generate haze-free image. Specifically, the disentangled content image is used as input for $G_C$ to generate clean images using adversarial and cyclic loss functions, and the output features of the haze encoder given an underwater image i.e., $F_{{h}_{w}}$ are used by $G_U$ for underwater regeneration. A detailed discussion of the restoration mechanism is given next.

\subsection{Restoration of underwater images}
\label{Sec:Restoration}
This section explains the methodology to restore underwater images using disentangled haze and content information from Sec. \ref{Sec:Haze_disentanglement}. The restoration mechanism consists of four networks, a Generator network $G_C$ to generate clean images, another Generator network $G_U$ to transfer images from clean to underwater domain, and two discriminator networks, $D_U$ and $D_C$, to differentiate between clean and underwater images. Below is a detailed discussion on different loss functions used in the restoration framework.

\textbf{Cycle consistency losses:} Adversarial losses alone are not sufficient to restore underwater images as different artifacts and unwanted colour shifts \cite{CycleGAN} can be observed in the generated clean images. To mitigate these effects, we use cycle consistency loss to ensure that the generated images are free from artifacts and remain faithful to the input images. 
% Motivated by CycleGAN, we use cycle consistency loss functions to ensure that generated images are artifact-free. To achieve this, the generated image $G_C(I_{con})$ is passed through $G_{U}$ to match the input image $I_{w}$. 
However, as pointed earlier in the introduction, generating an underwater image from a clean image is one-to-many mapping, i.e., $G_{U}$ can generate an underwater image with random haze information that does not match the input image which results in cycle consistency mismatch. We propose to solve this by using disentangled haze information from input underwater images. Specifically, given an underwater image $I^{\dagger}_{w}$, we know that HDN network disentangles haze information  $F_{h_{w}}$, using loss function $L_{d}$. The disentangled haze information $F_{h_{w}}$ is used in $G_U$ to generate an underwater image that matches the input image. The resultant cycle consistency loss is written as 
% \begin{equation}
% \begin{split}
%  {L}^{res}_{h_{cyc}}(G_C,G_U) =
%  ||G_U(G_C(I_{w}), F_{h_{w}}) - I_{w}||_1
% \end{split}
% \end{equation}
\begin{equation}
\begin{split}
 {L}_{r_3}(G_C,G_U) & =
 ||G_U(G_C(I^{\dagger}_{con}) + I^{\dagger}_{con}, F_{h_{w}}) - I^{\dagger}_{w}||_1 \\
  & = ||\hat{I}^{\dagger}_{w} - I^{\dagger}_{w}||_1 \\
\end{split}
\end{equation}
Similarly, cycle consistency loss for the clean image $I_c$ can be written as 
% \begin{equation}
% \begin{split}
%  {L}^{res}_{hf_{cyc}}(G_C,G_U) =
%  ||G_C(G_U(I_{c}, F_{h_{w}}) - I_{c}||_1
% \end{split}
% \end{equation}
% \begin{equation}
% \begin{split}
%  {L}_{r_4}(G_C,G_U) & = ||G_C(G_U(I^{\star}_{c}, F_{h_{w}})) - I^{\star}_{c}||_1\\
%   & = ||\hat{I}^{\star}_{c} - I^{\star}_{c}||_1 \\
% \end{split}
% \end{equation}

\begin{equation}
\begin{split}
 {L}_{r_4}(G_C,G_U) & = ||G_C(I^{\star}_{con}) + I^{\star}_{con} - I^{\star}_{c}||_1\\
  & = ||\hat{I}^{\star}_{c} - I^{\star}_{c}||_1 \\
\end{split}
\end{equation}

% We additionally use an SSIM \cite{ssimloss} loss to maintain the contents between the input and restored image. The loss can be written as 
% \begin{equation}
% \begin{split}
%  {L}_{r_5}(G_C) =
%  1 - \frac{1}{N} \sum_{n=1}^{N}(SSIM(n))
% \end{split}
% \end{equation}
% where N is the number of pixels in an image.

The final loss function to restore an underwater image is a combination of all the four loss functions
and can be written as 
% \begin{equation}
% \begin{split}
% L_{res} (G_C,G_U,D_U, D_C)  = \omega _1 {L}^{res}_{hf_{adv}} +  \\  \omega _2 {L}^{res}_{h_{adv}} + \omega _3 {L}^{res}_{h_{cyc}} + \omega _4  {L}^{res}_{hf_{cyc}}
% \end{split}
% \end{equation}
\begin{equation}
\label{Eq:restoration_losses}
\begin{split}
L_{r} (G_C,G_U,D_U, D_C)  = \omega _1 {L}_{r_1} +   \omega _2 {L}_{r_2} + \\ \omega _3 {L}_{r_3} + \omega _4  {L}_{r_4}% +  \omega _5  {L}_{r_5}
\end{split}
\end{equation}
where $\omega_1$, $\omega_2$, $\omega_3$ and $\omega_4$ are trade-off weights.

The total loss function to train HDN and restoration modules is the combination of Eq. \ref{Eq:disentanglement_losses} and 
Eq. \ref{Eq:restoration_losses} and can be written as
\begin{equation}
\label{Eq: total_loss}
    L_{total} = L_{d} + L_{r}
\end{equation}
where $L_{total}$ is the total loss function, $L_{d}$, and $L_{r}$ are disentanglement (Eq. \ref{Eq:disentanglement_losses})  and restoration (Eq. \ref{Eq:restoration_losses}) losses , respectively. 

\textbf{Testing:} HDN, $G_C$, $G_U$, $D_C$, and $D_U$ networks are trained in an end-to-end manner until convergence using Eq. \ref{Eq: total_loss}. A detailed discussion on experimental set-up and hyper-parameters is given in following section. To restore an underwater image ($I_w$) during test time, $I_w$ is first passed through the HDN network for content image $I_{con}$ (Eq. \ref{Eq. contentimage}). The resultant content image $I_{con}$ is passed through $G_C$ for the final restored image i.e.,
\begin{equation}
    % I_{res} = G_{C}(D(E_{hf}(I_w)))
     I_{res} = G_{C}(I_{con}) + I_{con}
\end{equation}
where $I_{res}$ is the final restored image.
\section{Experiments}
\label{Sec: Experiments}
This section is arranged as follows: (i) Implementation details. (ii) Datasets and metrics used. (iii) Ablation studies. (iv) Comparison results.
\subsection{Implementation details}
We used NVIDIA-2080 Ti GPU and Pytorch library to train and test our network. HDN and restoration modules are trained in an end-to-end manner using Eq. \ref{Eq: total_loss}. For Eq. \ref{Eq:disentanglement_losses} (HDN network), we empirically found $\lambda_1=1$,  $\lambda_2=10$, $\lambda_3=1$, and $\lambda_4=1$ to give best results and for Eq. \ref{Eq:restoration_losses}, we followed CycleGAN \cite{CycleGAN} with $\omega_1=1$, $\omega_2=1$, $\omega_3=10$, and $\omega_4=10$. We observed that the training converged in around 60-70 epochs and further continued training till 80 epochs. Following options are used to train our network: patch size of 128 and batch size of 4, ADAM optimizer for updating the weights, momentum values for optimizer with $\beta_1$ = 0.9, and $\beta_2$ = 0.99 and learning rate of 0.0005. Additional details about the training mechanism and architecture of the HDN and restoration modules are given in the supplementary material.

\subsection{Datasets and metrics}
\textbf{Datasets:} We used four publicly available UW datasets to train and evaluate our model: UFO-120 \cite{sesr}, UWNet \cite{euvp}, UWScenes \cite{euvp}, and UIEB \cite{uieb}. UFO-120 \cite{sesr} contains 1500 training samples and 120 testing samples. Underwater images in UFO-120 are generated using distortion mimicking CycleGAN \cite{CycleGAN} based model, followed by Gaussian blurring and bicubic interpolation. \cite{euvp} has two underwater image datasets, UWNet, and UWScenes, with corresponding ground truth images. UWNet has 3700 paired images for training and 1270 testing samples; UWScenes has 2185 training and 130 test images. The underwater images in UWNet and UWScenes are synthetically generated using a similar procedure as followed in \cite{sesr}. Recently, UIEB \cite{uieb} proposed a real underwater benchmark dataset with  paired images. The authors of \cite{uieb} collected different UW images from Google, Youtube and UW related works \cite{rcp,fusion-based,retinex,uieb1}. The reference images are carefully selected by human  volunteers from among the output of 12 enhancement methods, including nine underwater restoration methods, two dehazing methods and a commercial tool. In total UIEB \cite{uieb} has 890 UW images with corresponding references. Since the authors did not provide a train-test split, we randomly selected 800 images for training and the remaining 90 for testing. We observed that the UW images in UIEB are more challenging to restore and have a rich variety of scenes compared to other datasets.

\begin{figure*}[t]
%\newlength\fsdurthree
%\setlength{\fsdurthree}{-1.5mm}
\setlength{\tabcolsep}{1pt}
\scriptsize
\centering
%\hspace{-0.4cm}
\begin{tabular}{ccccccccccc}

\includegraphics[width=0.20\linewidth]{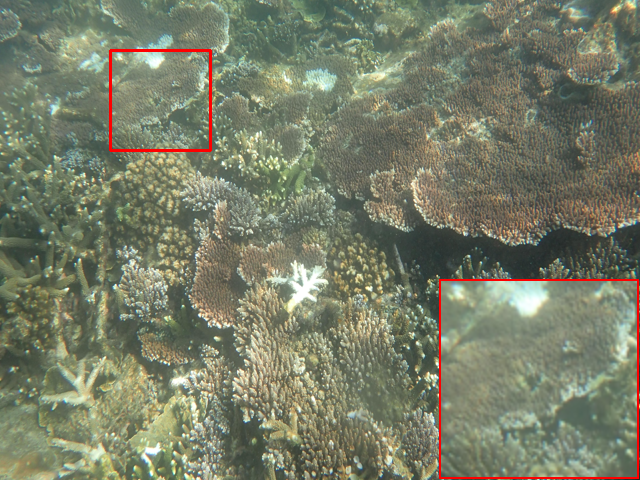} &
\includegraphics[width=0.20\linewidth]{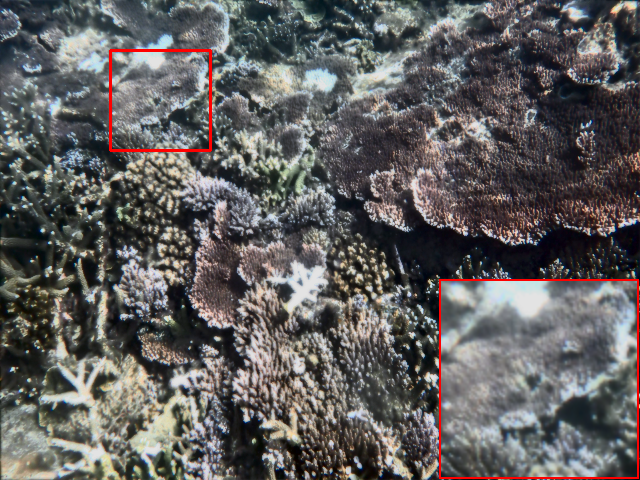} &
% \includegraphics[width=0.20\linewidth]{CycleGAN_882.png} &
% \includegraphics[width=0.20\linewidth]{UEGAN/882.png} &
%  \includegraphics[width=0.20\linewidth]{EnlightenGAN/882.png}\\
% Input & IBLA\cite{bluriness-based}  (TIP '17) & CGAN\cite{CycleGAN} (ICCV '17) & UEGAN\cite{UEGAN}  (TIP '20) &  EnGAN\cite{enlightengan} (TIP '21)\\
% \includegraphics[width=0.20\linewidth]{DCLGAN/882.png} &
% \includegraphics[width=0.20\linewidth]{ShadowGAN/882.png} &
\includegraphics[width=0.20\linewidth]{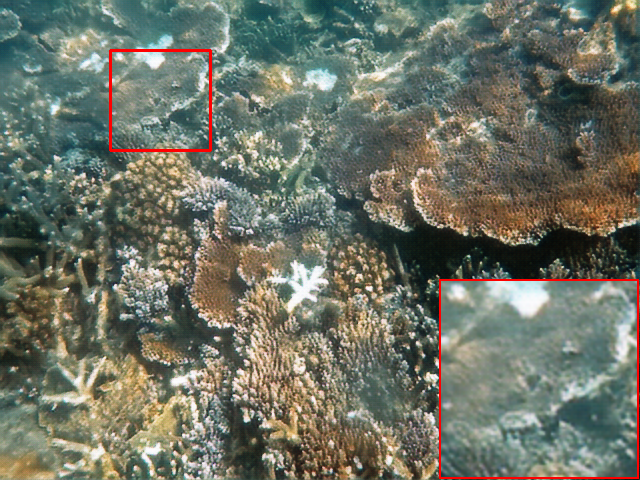}  &
\includegraphics[width=0.20\linewidth]{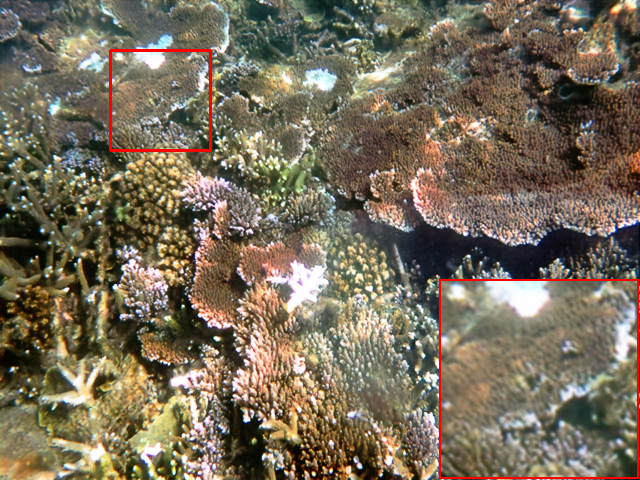}  &
\includegraphics[width=0.20\linewidth]{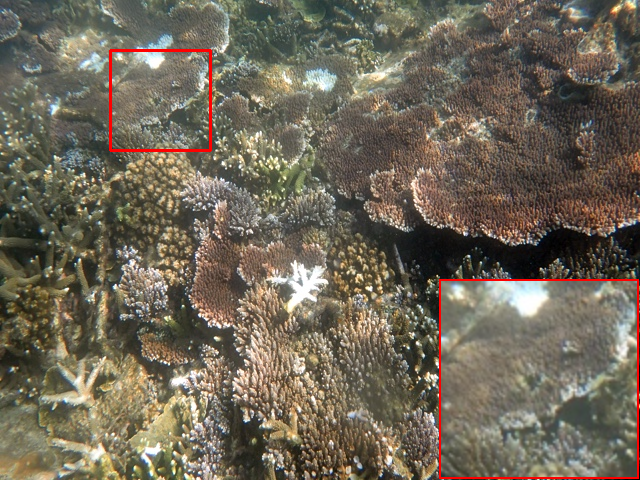}\\
Input & SGAN\cite{shadowgan} (ICCV '19) & UDSD\cite{chellapa} (CVPR '19)  & Ours.  & Ground truth.\\
% \includegraphics[width=0.20\linewidth]{Input/874.png} &
% \includegraphics[width=0.20\linewidth]{IBLA/874.png} &
% \includegraphics[width=0.20\linewidth]{CycleGAN/874.png} &
% \includegraphics[width=0.20\linewidth]{UEGAN/874.png} &
%  \includegraphics[width=0.20\linewidth]{EnlightenGAN/874.png}\\
% Input & IBLA\cite{bluriness-based}  (TIP '17) & CGAN\cite{CycleGAN} (ICCV '17) & UEGAN\cite{UEGAN}  (TIP '20) &  EnGAN\cite{enlightengan} (TIP '21)\\
% \includegraphics[width=0.20\linewidth]{DCLGAN/874.png} &
% \includegraphics[width=0.20\linewidth]{ShadowGAN/874.png} &
% \includegraphics[width=0.20\linewidth]{Chellapa/874.png}  &
% \includegraphics[width=0.20\linewidth]{Ours/874.png}  &
% \includegraphics[width=0.20\linewidth]{GT/874.png}\\
% DCLGAN\cite{contrastiveCycle} (ECCV '20) & SGAN\cite{shadowgan} (ICCV '19) & UDSD\cite{chellapa} (CVPR '19)  & Ours.  & Ground truth.\\
\end{tabular}\\ %\hspace{-2.3mm}

\caption{Visual comparisons on real underwater images from UIEB dataset \cite{uieb}. Zoomed portions of the selected regions are provided at the bottom right corner. Skip connection present in SGAN, UEGAN, EnGAN deteriorates the quality of generated images. CGAN, DCLGAN, and UDSD struggle to remove haze from challenging examples. In contrast, the results of our method generate almost haze-free images due to `consistent' cycle consistency and disentangled content information.}
\label{fig: Comparisons_uieb}
\end{figure*}

Since these datasets contain paired images, we used the following procedure to prepare the unpaired set. For every dataset with $X$ number of clean and haze paired images, $\frac{X}{2}$ haze images are randomly selected from the underwater set, and the corresponding ground-truth images are removed from the clean set. The remaining images in the clean set and the selected haze images are used for training. The same procedure is followed for all four datasets. This ensures that there are no paired images in the training set.

\subsection{Conclusions}
We proposed an unsupervised haze removal algorithm from underwater images using a haze disentanglement network (HDN) and a restoration module. HDN is used for disentangling haze and content from UW images. While the disentangled content is used as input for restoration module, the haze information is used for `consistent' cycle consistency. Different ablation studies revealed that the proposed HDN network successfully decouples haze and content from an underwater image. Comparisons with prior art show that our methodology visually improves  Visual comparisons onthe state-of-the-art and the quantitative metrics as well. We believe that the loss functions and the network architecture proposed in this paper will help improve the performance of unsupervised networks futher.

%  Visual comparisons on{Tex_files/proposed_method2}
% \input{Tex_files/comparisons_1}

% \input{Tex_files/data_augmentation}

% \input{Tex_files/Experiments}

% \input{Tex_files/table}

{\small
	\bibliographystyle{IEEEtran}
	\bibliography{egbib}
}
\end{document}